\documentclass{article}
\usepackage[utf8]{inputenc}
\usepackage{fancybox}
\usepackage{amsmath,amssymb,cite}
\usepackage{graphicx,times,multirow,grffile}
\usepackage{subfigure}
\usepackage[T1]{fontenc}
\usepackage{caption}
\usepackage{bm}
\usepackage{fullpage}
\usepackage[table,xcdraw]{xcolor}
\usepackage{authblk}
\usepackage{hyperref}

\title{Toward Interpretability of Dual-Encoder Models for Dialogue Response Suggestions }
\author[1]{Yitong Li}
\author[2]{Dianqi Li}
\author[3]{Sushant Prakash}
\author[3]{Peng Wang}
\affil[1]{Duke University, (yitong.li@duke.edu)}
\affil[2]{University of Washington, (dianqili@uw.edu)}
\affil[3]{Google Research (sush@google.com, wangpengatwork@gmail.com)}
\date{September 2019}

\begin{document}
\maketitle

\begin{abstract}
This work shows how to improve and interpret the commonly used dual encoder model for response suggestion in dialogue. We present an attentive dual encoder model that includes an attention mechanism on top of the extracted word-level features from two encoders, one for context and one for label respectively. To improve the interpretability in the dual encoder models, we design a novel regularization loss to minimize the mutual information between unimportant words and desired labels, in addition to the original attention method, so that important words are emphasized while unimportant words are de-emphasized. This can help not only with model interpretability, but can also further improve model accuracy. We propose an approximation method that uses a neural network to calculate the mutual information. Furthermore, by adding a residual layer between raw word embeddings and the final encoded context feature, word-level interpretability is preserved at the final prediction of the model.
We compare the proposed model with existing methods for the dialogue response task on two public datasets (Persona and Ubuntu). The experiments demonstrate the effectiveness of the proposed model in terms of better Recall@1 accuracy and visualized interpretability.
\end{abstract}

\section{Introduction}\label{sec:introduction}

Deep learning based dialogue systems have shown promising performance in many applications such as smart reply~\cite{henderson2017efficient}, conversation semantic embedding~\cite{yang2018learning}, human-computer interaction~\cite{ghazvininejad2018knowledge} and others ~\cite{mazare2018training,zhang2018personalizing,zhao2017learning}. Deep neural nets extract rich representations with high-level semantic information that are useful for message retrieval~\cite{yang2018learning,xu2019gated} and response generation~\cite{zhao2017learning} in conversations.

The dual encoder model~\cite{henderson2017efficient,yang2018learning} is widely used among various dialogue  models especially for retrieving response messages, due to its simple structure and competitive computational speed. The dual encoder model consists of two separate encoders, which extract features for the dialogue context (e.g. the previous messages) and a candidate response, respectively. Then a similarity score is computed between the extracted context and candidate features. Whichever candidate from a predefined list has the highest score is selected as the best response. Such a "retrieval" based method shows many advantages over generative language models~\cite{zhao2017learning} in industrial applications, such as computational efficiency and preventing undesirable responses. 

In this work, we focus on interpreting and improving the dual encoder model, which is normally considered a black box. Although there are many existing models with interpretability designed for question answering~\cite{serrano2019attention_interpretable,palangi2018question,trott2017interpretable,agrawal2018don} or textual entailment~\cite{lin2017structured,rocktaschel2015reasoning,zhao2016textual}, fewer works have investigated interpretability in dialogue response generation. In this paper, we present an attentive dual encoder model, which adds an attention mechanism on top of the extracted word-level features from both encoders. With this pairwise word-level attention, not only is the prediction accuracy improved, but also the most important context and response words contributing to the decision of the model can be highlighted. Interpreting a model in terms of the relationship between inputs and outputs can greatly assist developers to debug and improve models, and help users understand why a certain result is suggested.

\begin{figure}
    \centering
    \includegraphics[width = 0.6\textwidth]{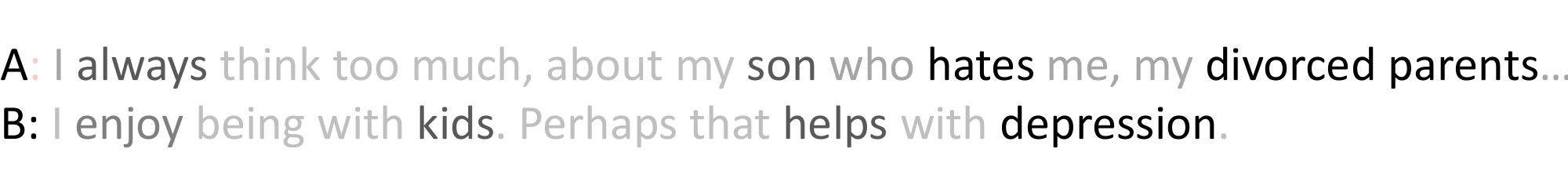}
    \caption{Attention learned without constraining on unimportant words. Words with higher learned wights are highlighted in darker color. 
    This example is selected from Persona dataset. \textbf{A} and \textbf{B} denote two different people in one conversation.~\cite{zhang2018personalizing}}
    \label{fig:intuition_example}
\end{figure}
There are two potential problems when directly applying the attention mechanism at the word-level. First, the standard attention mechanism only emphasizes predictive words to optimize the training loss, without any constraints on attentions weights for the purpose of interpretation. The example in Figure~\ref{fig:intuition_example} shows that many unimportant words are highlighted
, such as `about' and `perhaps'. Emphasizing unimportant words muddles interpretability and may also harm model performance by over-fitting to the training set. The second problem is caused by commonly used text encoder structures. Most existing text encoders, such as LSTM and Transformers, discard fine-grained word-level information and create representations that entangle information from across the whole sentence. While this brings advantages for sentence prediction tasks, it impedes word-level interpretation of the prediction.

In order to solve the first problem, we borrow an idea from information theory. Intuitively, prediction-related words should contain useful information while unimportant words should have little information for response retrieval. Thus, in addition to the original attention method, we design a novel regularization loss that minimizes the mutual information between unimportant words in the context and the desired response, so that important words are emphasized while unimportant words are de-emphasized. We propose an approximation method to calculate the mutual information by using a neural network. In practice, this loss can improve both the quality of interpretability as well as response retrieval accuracy.

For the second problem, we present a simple yet effective solution that uses a residual layer connecting raw word embeddings and the final encoded context feature. By tuning the weights on the raw word embeddings, we can balance the importance of the encoded contextual information (for retrieval accuracy) and individual word features (for interpretability at the word-level).

In summary, our contributions are three-fold:
\begin{itemize}
    \item Introduced a learnable attention mechanism between input dialogue context and response text pairs, which improves both retrieval accuracy and interpretability.
    \item Proposed a regularization term that emphasizes important word pairs and penalizes unimportant word pairs, therefore improving interpretability in an unsupervised way.    
    \item Demonstrated that the fusion of both encoded features and word specific embeddings further improves the interpretability.
\end{itemize}
\section{Related Works}\label{sec:related_works}

Traditional dialogue systems can be roughly divided as goal-oriented and non-goal driven models~\cite{zhang2018personalizing}. Goal-oriented models target specific application circumstances~\cite{young2013pomdp}, such as customer service~\cite{xu2017new} and computer system troubleshooting~\cite{lowe2015ubuntu} dialogues, etc. These works tend to use lexical semantics to match basic syntactic similarity~\cite{jimenez2012soft}. In contrast, non-goal driven models focus more on data statistics, rather than using hand-coded rules. They score responses based on how well they match the dialogue context.

With the development of neural networks, recent dialogue systems are usually non-goal driven and trained in an end-to-end fashion. Among them, generative dialogue systems have attracted growing interest in recent years~\cite{zhao2017learning,ghazvininejad2018knowledge,song2018towards}, especially among the research community. They are designed to learn the conditional distribution of the responses given dialogue  history~\cite{yan2018smarter,sordoni2015neural}. ~\cite{song2018towards} uses Determinantal Point Processes to generate responses with diversity and ground knowledge can also be utilized to generate novel results~\cite{ghazvininejad2018knowledge}. The most significant problem that hinders wide industrial use of generative conversation models is reliability. Most existing generative models suffer issues of incorrect grammar, lack of long-term coherence, and even generation of offensive responses~\cite{wallace2019universal}.

Compared with generative conversation models, dialogue retrieval based models are more reliable and simpler in structure~\cite{henderson2017efficient}. The well known dual encoder model~\cite{henderson2017efficient,yang2018learning} has had success with semantic similarity and response scoring in conversations. Recent works tackle more challenging situations, like multi-party conversation recommendation~\cite{zhang2018addressee}, multi-turn response selection~\cite{zhou2018multi}, and personalization~\cite{zhang2018personalizing,mazare2018training}.

Since most existing deep learning models are black boxes, interpretability becomes a desired property to explain why the neural network gives a certain result. Interpretable neural networks have been developed for text generation~\cite{yu2018qanet,zhao2018unsupervised}, visual question answering~\cite{vedantam2019probabilistic,anderson2018bottom} and sequential data classification~\cite{li2017targeting}.~\cite{liang2018focal} further extends single image question answering to a collection of images. Most existing works are based on generative models, where interpretability is usually achieved via a Variational AutoEncoder (VAE)~\cite{hu2017toward,hsu2017unsupervised,zhao2018unsupervised}. To further improve interpretability, an attention mechanism~\cite{vaswani2017attention} is integrated to most existing methods~\cite{sydorova2019interpretable,vedantam2019probabilistic,anderson2018bottom,niu2019recursive}. Although interpretability has been applied in the domains mentioned above, there are few works that aim to interpret neural conversation models~\cite{palangi2018question}. 


With the development of a technique to estimate mutual information (MI) in high dimensional data~\cite{belghazi2018mine,gabrie2018entropy}, using knowledge from information theory to improve the performance of neural networks has received growing attention. For instance, Deep Infomax~\cite{hjelm2018learning} learns generalized features for images by maximizing the MI between global and localized features.~\cite{li2019way} focuses on dynamic scene navigation with MI and ~\cite{qu2019weakly} applies MI on the graph aligning task. Mutual information has became an effective and efficient way to measure the correlation among random variables in neural networks.
 
\section{Model}\label{sec:model}

\subsection{Preliminary}
Given a dialogue context $\bm x$, which is sampled from a dialogue context set $\bm x \in \bm X$ and contains at most $T$ messages $\bm x = \{\bm x_1,\cdots, \bm x_T \}$, the response suggestion task is to retrieve the best response from a given response candidate list. Note that there could be other non-text input signals associated with each message, such as the user id. Such signals can be encoded in the same way as word embeddings. 
For simplicity, we leave out the user id signal in the problem formulation below.
The response candidate list contains 
messages $\bm Y = \{\bm y_1,\cdots, y_l,\cdots, \bm y_L \}$ for possible responses. $L$ is the total number of candidates and can be on the order of tens of thousands in real-world applications.  The best response message $\bm y \in \bm Y$, corresponds to the label of the dialogue context $\bm x$.

\begin{figure}[htb]
\centering
\subfigure[Baseline]{\label{fig:encoder_framework} 
\includegraphics[scale=0.34]{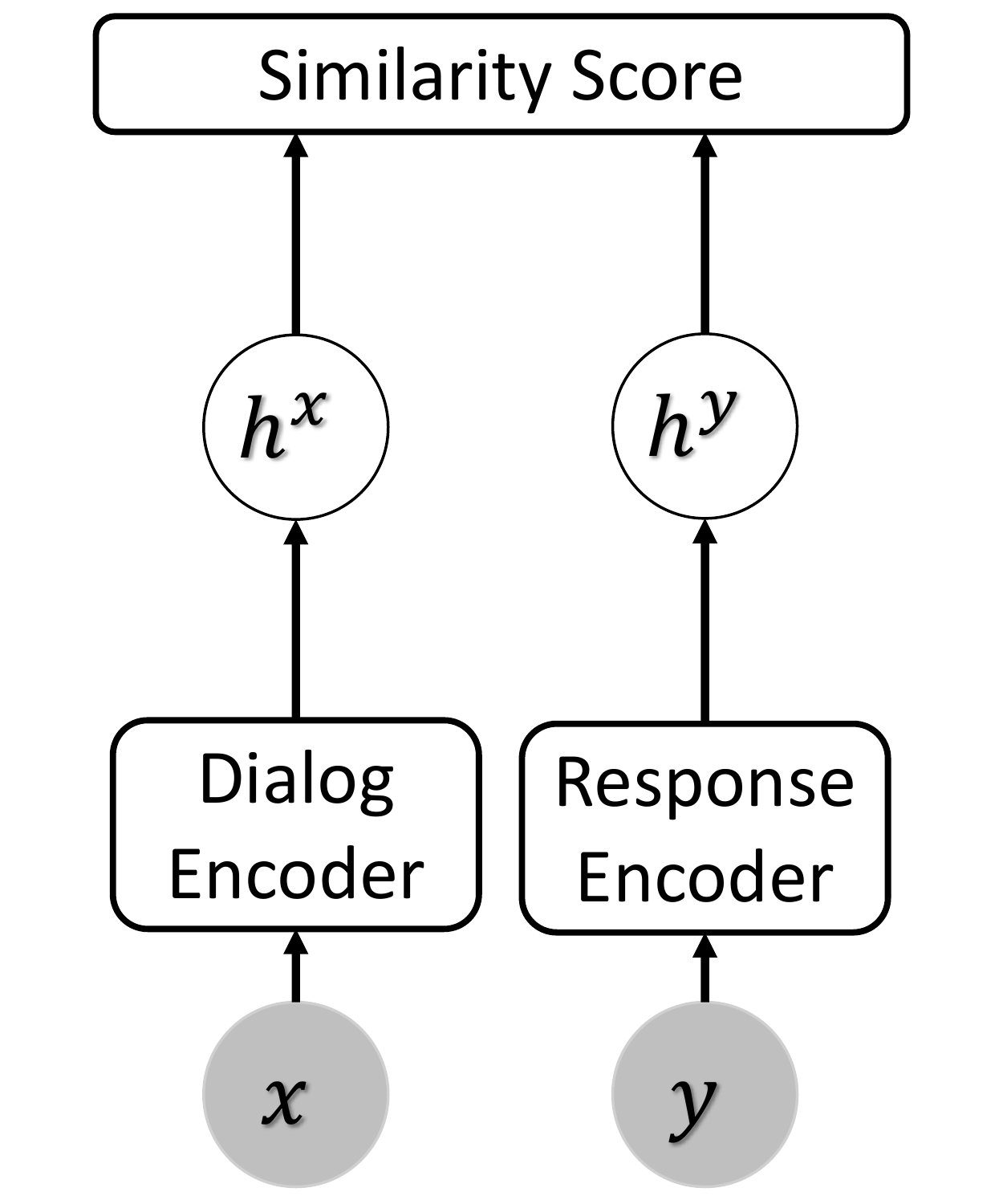}}
\subfigure[Attentive Dual Encoder Model]{\label{fig:attention_framework} 
\includegraphics[scale=0.34]{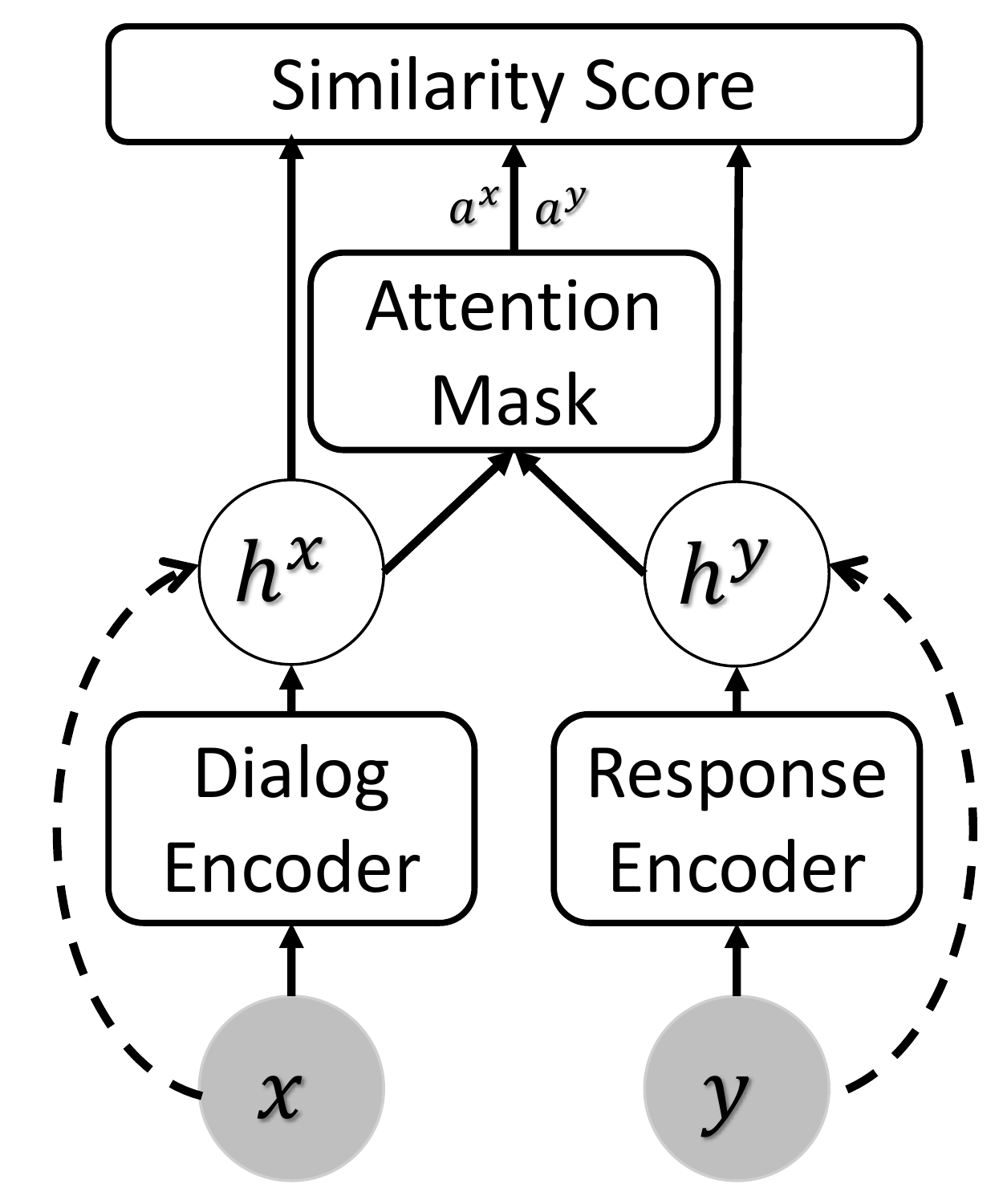}}
\caption{Comparison of dual encoder model and our attentive dual encoder model. 
}
\label{fig:samples_all}
\vspace{-3mm}
\end{figure}

The previous work~\cite{henderson2017efficient} defines an dual encoder model, where the features of dialogue $\bm x$ and each label $\bm y$ ($\bm y \in \bm Y$) are extracted by a dialogue encoder and a response encoder, respectively. The framework is given in Figure~\ref{fig:encoder_framework}. The two encoders can be designed with partially shared or totally separate structures based on the training size of $\bm X$ and $\bm Y$, while the word embedding is usually shared. Denote $\bm h^{\bm x}=E(\bm x; \bm \theta_{\bm x})$ and $\bm h^{\bm y} = E(\bm y; \bm \theta_{\bm y})$ the corresponding encoded token-level features of the dialogue context and response. $\bm h^{\bm x} \in \mathbb{R}^{n_x \times d}$ and $\bm h^{\bm y} \in \mathbb{R}^{n_y \times d}$ where $n_x$ and $n_y$ are the lengths of the dialogue context and response, respectively, and $d$ is the dimension of encoded tokens. The training objective is to maximize the similarity score of paired dialogue context-response samples $(\bm x, \bm y)$ while minimizing the scores from other mismatched pairs.

The similarity score can be formulated by $\cos(f(\bm h^{\bm x}), f(\bm h^{\bm y}))$, where $\cos$ denotes the cosine similarity function and $f$ is a function that aggregates the encoded token-level features $\bm h^{\bm x}, \bm h^{\bm y}$ into a fixed dimension $d^\prime$. Since the length of each sentence is varied, an average pooling function $f$ across the token length dimension can be used to get a final feature vector with fixed dimension $d^\prime = d$. Specifically, $f(\bm h^{\bm x}) \in \mathbb{R}^d$ and $f(\bm h^{\bm y}) \in \mathbb{R}^d$.
Although this method is simple and effective in practice, it is a black box without interpretability.

On top of the dual encoder model, we introduce a more interpretable model, called "attentive dual encoder model", in Section~\ref{subsec:attentive_encoder_encoder}. In Section~\ref{subsec:non-attention_mechanism}, we introduces a new loss term in the attentive dual encoder model to regularize the learning of the attention mask that emphasizes important tokens and de-emphasizes unimportant tokens. To improve interpretability at the encoded feature layer where word-level information is entangled with contextual information, in Section~\ref{subsec:combine_word_embedding}, 
we propose a residual layer that leverages the raw word embedding.

\subsection{Attentive dual encoder Model}\label{subsec:attentive_encoder_encoder}
An attentive dual encoder model is introduced to learn the connection between dialogue  context and response at the word-level. We adopt the attention mechanism~\cite{vaswani2017attention} on top of the standard dual encoder model, as shown in Figure~\ref{fig:attention_framework}. Specifically, a similarity matrix $\bm S \in \mathbb{R}^{n_x \times n_y}$ is defined on the encoded features $\bm h^{\bm x}$ and $\bm h^{\bm y}$ to measure the pairwise word relationships. The $ij$-th entry of $\bm S$ is given as 
\begin{equation}\label{eq:Sij}
    S_{ij} = \text{sim}(\bm h^{\bm x}_i, \bm h^{\bm y}_j); \hspace{2mm} 1 \leq i \leq n_x, 1 \leq j \leq n_y,
\end{equation}
where $\bm h^{\bm x}_i$ is the $i$th word feature of the dialogue and $\bm h^{\bm y}_j$ is the $j$th word feature of the response. For simplicity, the similarity function $\text{sim}(\cdot)$ is the cosine similarity~\cite{faghri2017vse++} though other similarity functions~\cite{shen2018disan} are also applicable. 

Given $\bm S$, the intuition is to find out if two words have strong connections, i.e. if there are any words in the dialogue context and responses that greatly influence the prediction. In the final response prediction, the dialogue context and response features are weighted according to the similarity matrix. For each word in the dialogue context, we first select the words from the candidate response with the maximum similarity, and vice versa for each word in the candidate response.
Specifically, the maximum pooled attention weight for dialogue context and response can be defined as
\begin{equation}\label{eq:define_a}
    a^{\bm x}_i = \max_j \frac{e^{S_{ij}}}{\sum_{j} e^{S_{ij}}},\ a^{\bm y}_j = \max_i \frac{e^{S_{ij}}}{\sum_{i} e^{S_{ij}}} ,
\end{equation}
where $i$ and $j$ are the indexes for word in dialogue  and response, respectively. Then the final attention weight for dialogue context and response are defined as $\bm a^{\bm x} = softmax(a^{\bm x}_1,\cdots,a^{\bm x}_{n_x})$ and $\bm a^{\bm y} = softmax(a^{\bm y}_1,\cdots, a^{\bm y}_{n_y})$.
Note that other attention mechanisms can also be adopted, like mean pooling or weighted mean pooling w.r.t the similarity matrix $\bm S$~\cite{faghri2017vse++}. 

The attention weights the original encoded feature $\bm h^{\bm x}$ and $\bm h^{\bm y}$ by their importance as $(\bm a^{\bm x})^{\intercal} \bm h^{\bm x}$ and $(\bm a^{\bm y})^{\intercal} \bm h^{\bm y}$. The final prediction score of dialogue context-response pair is given as
 \begin{equation}\label{eq:attentive_prediction_score}
 score(\bm x, \bm y) =   dot((\bm a^{\bm x})^{\intercal} \bm h^{\bm x}, (\bm a^{\bm y})^{\intercal} \bm h^{\bm y} ) ,
\end{equation}
where the dot product can be replaced by other metrics~\cite{wang2019multi}. 

To train the model, the observed pairs of context and response $(\bm x, \bm y^{\bm x})$ are considered positive pairs and should have higher scores, while all other mismatched pairs of context and response $(\bm x, \bm y^{\bm x^\prime})$, where $\bm x^\prime \neq \bm x$, are negative pairs that should have lower scores. However, randomly sampling negative training pairs is time consuming. Practically, we construct a mini-batch $\bm D = \{(\bm x^i, \bm y^i)\}^N_{i = 1}$ by sampling $N$ positive dialogue-response pairs, where all mismatched pairs $(\bm x^i, \bm y^j), j \neq i$ are used as negative pairs. We therefore conduct the retrieval task as a dialogue context-response matching problem in each mini-batch. A softmax retrieval loss can be defined as:
\begin{equation}
    \mathcal{L}_{ret, \bm y} = - \sum^{N}_{i = 1}log\frac{exp(score(\bm x^i, \bm y^i)/\gamma)}{\sum^{n}_{j = 1} exp(score(\bm x^i, \bm y^j)/\gamma)}, 
\end{equation}{}
where $\gamma$ is a temperature parameter that normalizes the context-response similarity to a proper range. Since in each mini-batch, dialogue contexts and responses for dual encoder modeling are symmetric, we can also use each response to retrieve its corresponding context. Therefore, a response-context retrieval loss can be written as:
\begin{equation}
    \mathcal{L}_{ret, \bm x} = - \sum^{N}_{i = 1}log\frac{exp(score(\bm x^i, \bm y^i)/\gamma)}{\sum^{n}_{j = 1} exp(score(\bm x^j, \bm y^i)/\gamma)}.
\end{equation}{}
The overall retrieval loss for the proposed attentive dual encoder is:
\begin{equation}
    \mathcal{L}_{ret} = \mathcal{L}_{ret, \bm x} + \mathcal{L}_{ret, \bm y}
\end{equation}{}



\subsection{Non-attention Regularization}\label{subsec:non-attention_mechanism}

As illustrated in Figure~\ref{fig:intuition_example}, it is possible that the learned attention has higher weights on unrelated information due to limited training samples or biased words with high frequency in the training set. As a result, the learned attention can be noisy for the purpose of interpretation. In this section, we  introduce a non-attention regularization mechanism 
to help the model attend on semantically important words, while ignoring unimportant words. 

Recall that the attended dialogue context feature is $(\bm a^{ \bm  x})^{\intercal}\bm h^{\bm  x}$, where all entries in the attention weights are positive and have summation of one. In contrast to attention weights, we define $\bm 1 - \bm a^{\bm  x}$ as the non-attention weight, which means the model should de-emphasize those non-attended words during prediction. By applying the non-attended weight on the encoded features, the unimportant feature for dialogue context is defined as 
\begin{equation}\label{eq:non_attention_encodings}
\bar{\bm h}^{\bm x} = \left(\bm 1 - \bm a^{\bm x} \right)^{\intercal} \bm h^{\bm x} .
\end{equation}
Analogously, we can derive $\bar{\bm h}^{\bm y} = \left(\bm 1 - \bm a^{\bm y} \right)^{\intercal} \bm h^{\bm y}$ for the unimportant response feature. 

Ideally, $\bar{\bm h}_{x}$ should contain little information about the response. In order to achieve this, we adopt mutual information $I(\cdot)$ from information theory as the evaluation metric. In our situation, we use it to measure the uncertainty of the correct response $\bm y$ given the unattended dialogue context feature. Thus the new  mutual information is used as a regularization objective and can be written as $\min_{\bm \theta_{\bm x}, \bm \theta_{\bm y}} I(\bar{\bm h}^{\bm x}; \bm h^{\bm y})$.

However, it is not straightforward to calculate $I(\bar{\bm h}^{\bm x}; \bm h^{y})$ in a high-dimensional space. Inspired by the recent work~\cite{belghazi2018mine,poole2019variational}, we adopt a neural network to approximate this mutual information value. From ~\cite{donsker1975asymptotic}, $I(\bar{\bm h}^{x}; \bm h^{\bm y})$ can be upper bounded by the following formulation, given samples $\{\bm x_n, \bm y_n\}$, $n=1,\cdots,K$ in one mini-batch.
\begin{equation}\label{eq:reg_mutual_info}
    I(\bar{\bm h}^x; \bm h^y) \leq \mathbb{E}\left[ \frac{1}{K} \sum_{n=1}^K \left[ \log \frac{p(\bm h^y_n|\bm h^x_n)}{\frac{1}{K-1}\sum_{n\neq n'} p(\bm h^y_n | \bm h^x_{n'})} \right] \right] ,
\end{equation}
where $n$ and $n'$ are used as index of training samples. The expectation is taken over $\prod_n p(\bm h^x_n, \bm h^y_n)$. 

In practice, this is similar to the discriminator in a Generative Adversarial Network (GAN)~\cite{goodfellow2014generative}. Specifically, $p(\bm h^y_n| \bm h^x_{n})$ in Eq.~\eqref{eq:reg_mutual_info} contains correct sample pairs (real samples) while $p(\bm h^y_n| \bm h^x_{n'})$ ($n\neq n'$) is the distribution of mis-matched pairs (fake samples). In the following, we suppose $p(\bm h^y_n| \bm h^x_{n'})$ is approximated by a neural network with parameter $\bm \theta_D$. This network classifies true or false pairs from the given $\bm h^x$ and $\bm h^y$. Note that $p(\bm h^y_n| \bm h^x_{n'})$ can also be simplified as the vector inner product without any learnable parameters. When updating ~\eqref{eq:reg_mutual_info}, we use moving average for each mini-batch to alleviate the biased gradient problem. Further discussion of this point can be found in~\cite{belghazi2018mine,poole2019variational}.

\subsection{Combine Word Embeddings}\label{subsec:combine_word_embedding}
As pointed out by~\cite{gino2019validity,serrano2019attention_interpretable}, using only the features computed after attention can lead to inaccuracies. The output of the encoder can mix representations of multiple words, even the whole sentence, depending on the encoder structure. For a standard transformer model~\cite{vaswani2017attention}, encoded word-level features after the first layer only have $50\%$ of the information of the original word embedding. After ten layers, this number drops down to $7.5\%$. Therefore, the attention weights $\bm a^{\bm x}$ and $\bm a^{\bm y}$, calculated from the deeply encoded features $\bm h^{\bm x}$ and $\bm h^{\bm y}$, have a smoothed distribution at the word-level, undermining the interpretability of word importance during the prediction.


We use a simple yet effective method to address this issue, by adding raw word embeddings directly to the encoded features after multiple layers. As illustrated in Figure~\ref{fig:intuition_example}, this can be done by a residual layer between the raw word embeddings and the top layer of the dialogue context or response encoder. Taking the dialogue context $\bm x$ as example, the residual feature learned by the raw word embedding can be written as:
\begin{equation}\label{eq:combine_embedding}
\bm r^{\bm x} = \mathcal{F}\left(\bm e^{\bm x} ; \bm \theta_{\bm x} \right) ,
\end{equation}
where $\bm e^{\bm x}$ contains raw word embeddings for each word in dialogue  $\bm x$ in each column. To simplify the notation, parameters in $\mathcal{F}(\cdot)$ are included in the pre-defined parameter sets $\bm \theta_{\bm x}$ or $\bm \theta_{\bm y}$. 
$\mathcal{F}(\cdot)$ is implemented by a single fully connected layer in the experiment, which ensures that $\bm h^{\bm x}$ and $\bm r^{\bm x}$ are of the same dimension. $\bm r^{\bm x}$ can be concatenated or directly added to $\bm h^{\bm x}$. In this work, the final word embedding is calculated as $\alpha \bm h^{\bm x} + (1-\alpha) \bm r^{\bm x}$, where $\alpha$ is 
determined by the validation set. Thus the effect of individual word information is explicitly considered in the final encoded feature representation. In the experiment section, applying Eq.~\eqref{eq:combine_embedding} allows us to better discriminate the importance of individual words when visualizing the attention.

\subsection{Train the Attentive dual encoder Model}\label{subsec:model_overview}

The overall training objective of the attentive dual encoder model is given as 
\begin{equation}\label{eq:training_loss}
    \min_{\theta_x, \theta_y}\max_{\theta_D}\mathcal{L}_{ret}(\bm \theta_{\bm x}, \bm \theta_{\bm y}) +  \beta \mathcal{L}_{reg}(\bm \theta_{\bm x}, \bm \theta_{\bm y}, \bm \theta_D),
\end{equation}
where $\beta$ is a hyper-parameter to balance the value of regularization term. During the experiment, $\beta$ is set to 1 based on validation set. The retrieval cost $\mathcal{L}_{ret}$ is used to maximize the score of correct dialogue context-response pairs within one mini-batch, while the mutual information regularization term $\mathcal{L}_{reg} $ is used to force the attention weights to highlight useful information only.

 The training objective is a min-max game between the dual-encoder and the neural mutual information estimator. Note that parameter $\bm \theta_D$, which is used to estimate the mutual information, only appears in the second term of Eq.~\eqref{eq:training_loss}. 
 The update of $\bm \theta_D$ can be separated from the cross-entropy loss, while the update of $\bm \theta_{\bm x}$ and $\bm \theta_{\bm y}$ need to consider gradients from both terms.

\section{Experiment}\label{sec:experiment}

\paragraph{Experimental Setup:} Both dialogue context and response encoders are built upon Transformer~\cite{vaswani2017attention} with three layers. The dimensionality of the embedding and the number of head are set to 128 and 4, respectively. The word embedding dimension is set to 100. We used the Adam optimizer~\cite{kingma2014adam} with a learning rate $10^{-4}$, and a batch size of 64.
We also conduct an ablation study to show the effectiveness of each proposed component in our network. In the ablation study, \textbf{DE} denotes the standard dual encoder (DE) model~\cite{henderson2017efficient}. \textbf{ADE} is the Attentive dual encoder model with the additional attention mechanism introduced in Section~\ref{subsec:attentive_encoder_encoder}. \textbf{WE} is the acronym for Word Embedding, where a residual layer is connected between the raw word embedding and the output encoded features. \textbf{REG} represents the mutual information regularization term introduced in Section~\ref{subsec:non-attention_mechanism}.

\paragraph{Baseline.} The model is compared with several existing works. IR baseline~\cite{sordoni2015neural} measures the TF-IDF weighted cosine similarity between the bags of word features of dialogue  and messages in the candidate list. Similarly, Starspace~\cite{wu2018starspace} is trained by maximizing the learnable word embeddings between dialogue contexts and responses. Both of these two methods do not have any text encoder involved. KVPM~\cite{zhang2018personalizing} uses memory network and performs attention over dialogue contexts. It was originally designed for personalized dialogue model, while it is used without the user profile information in the experiment. 

\paragraph{Dataset:} In the experiment, we evaluate the proposed attentive dual encoder model with existing methods on two public datasets:

\textit{Ubuntu Dataset}~\cite{lowe2015ubuntu} contains $1,422,295$ training dialogues. The testing set contains $370, 573$ dialogues. Each dialogue contains four to five utterances. Since most of the response messages appear at a low frequency, only the top $1,000$ most common messages are selected for the response candidate list. Only dialogues with responses included in this list are evaluated.

\textit{Persona Dataset}~\cite{zhang2018personalizing} was initially published for developing personalized dialogue agents. It contains a total of $162, 064$ utterances over $10, 907$ dialogues, where $1,000$ dialogues with $15, 602$ utterances are used as training and $968$ dialogues with $15, 024$ utterances as testing. Instead of using a fixed candidate list, for each test dialogue, we randomly sample $19$ responses from other dialogues and combine with the ground truth response to evaluate scoring of the 20 candidates. 

\paragraph{Evaluation.} In the testing stage, we rank candidate responses from a candidate list using the score between the candidate and dialogue context (Eq.~\eqref{eq:attentive_prediction_score}). In the experiment, we only use Recall@1 as the quantitative evaluation metric, which matches real usages of the model. Recall@k is the accuracy defined as
\begin{equation}\label{eq:recall_k}
\text{Recall@k} = \frac{\sum^N_{n = 1}\textbf{I}\left( R_{\bm x_n,  \bm y_n} ,  k \right)}{N} ,
\end{equation}
where $N$ is the number of total evaluated instances. $R_{\bm x_n,  \bm y_n}$ is the rank of the similarity score between the dialogue context $\bm x_n$ and its ground truth response $\bm y_n$ in the final sorted list, and
\begin{equation}
\textbf{I}\left( R_{\bm x_n,  \bm y_n}, k \right) =\left\{
\begin{array}{ll}
                  1, ~~~R_{\bm x_n,  \bm y_n} \leq k\\
                  0,  ~~~R_{\bm x_n,  \bm y_n} > k\\
                \end{array}
                 \right.
\end{equation}

For the Ubuntu dataset, we also use the prior knowledge of response frequencies to further remove noise caused by rarely used responses. 
Denote $f_{\bm y_l}$ as the normalized usage frequency of response $\bm y_l$, $l=1,\cdots,L$ in the training set. Then the prediction score with prior knowledge can be written as 
\begin{equation}\label{eq:prior_prediction_score}
    score^\prime(\bm x_n, \bm y_l) = \log f_{\bm y_l} + score(\bm x_n, \bm y_l) ,
\end{equation}
which is computed for each message $\bm y_l$ in the candidate list.

\subsection{Quantitative Results}\label{subsec:exp_recall}
\begin{table}[htb]
  \centering
  \begin{tabular}{l|c|c}
  \hline
  & Persona & Ubuntu \\
  \hline 
   IR baseline~\cite{sordoni2015neural} & 24.1 & N/A \\
   Starspace~\cite{wu2018starspace} & 31.8 & N/A \\
   KVPM~\cite{zhang2018personalizing} & 34.9 & N/A \\
   \hline
    DE & 35.2 &  7.6\\
    ADE & 35.8 & 15.9  \\
    ADE + REG & 38.0 & 16.0 \\
    ADE + WE & 36.2 & 15.3  \\
    ADE + WE + REG & 38.1 & 15.6 \\
    \hline
  \end{tabular}
  \caption{Recall@1 in percentage comparison on two dialogue datasets. \label{tab:persona_result}}
\end{table}

We summarize the evluated results from the two datasets in Table~\ref{tab:persona_result}. As shown, the attentive dual encoder (ADE) outperforms other baselines on Recall@1 accuracy. Since neither the IR baseline~\cite{sordoni2015neural} nor Starspace~\cite{wu2018starspace} has an effective text encoder, their results are not competitive with others. Though KVPM~\cite{zhang2018personalizing} shares the text encoder for the dialogue context and the response, it does not have the non-attention mechanism. As a result, KVPM is not able to accurately select a response from a large candidate list. 

We also compare the model in an ablation study, where the mutual information regularization term and word embedding layers are gradually added to the model. As can be seen, the dual encoder (\textbf{DE}) has the lowest result, while it is still higher than other baselines. Adding the attention mechanism improves the model performance on Persona dataset, while there is a large performance gain on Ubuntu. This demonstrates that the attentive dual encoder model can not only help with the visualization, but can also improve the retrieval accuracy.

Adding the mutual information regularization term can further improve the results of the attentive dual encoder model. This contributes to the fact that irrelevant words are excluded by the explicit constraint of mutual information regularization. This not only alleviates the overfitting, but also helps the visualization for interpretation. Adding the word embedding residual layer (\textbf{WE}) does not have significant difference. This is reasonable since most of the textual knowledge has already been encoded in the feature vectors. However, it can help with the attention visualization, which will be shown in the next section. 
Additionally, compared with the parameters in the \textbf{DE} model, the parameters in the attention (Eq.~\eqref{eq:Sij}) and \textbf{WE} components (Eq.~\eqref{eq:combine_embedding}) are negligible. 

\subsection{Attention Visualization}\label{subsec:exp_visualization}

\begin{figure}[h!]
\centering
\subfigure[Attentive dual encoder]{
\includegraphics[width=.8\textwidth]{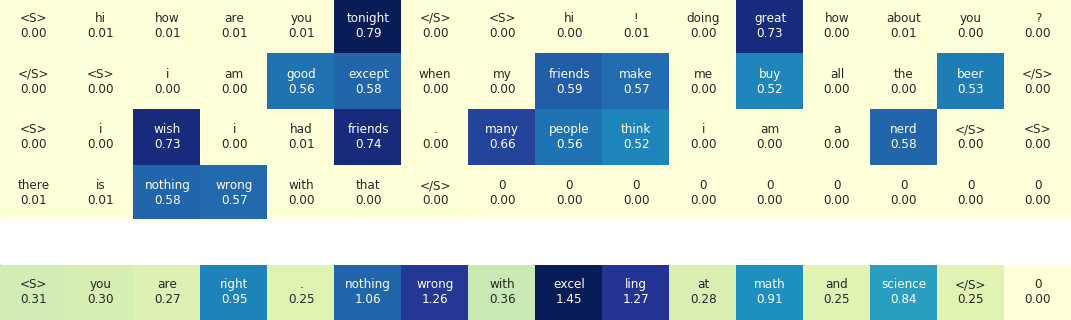}\label{fig:persona_vi1}}
\subfigure[ Attentive dual encoder + WE]{
\includegraphics[width=.8\textwidth]{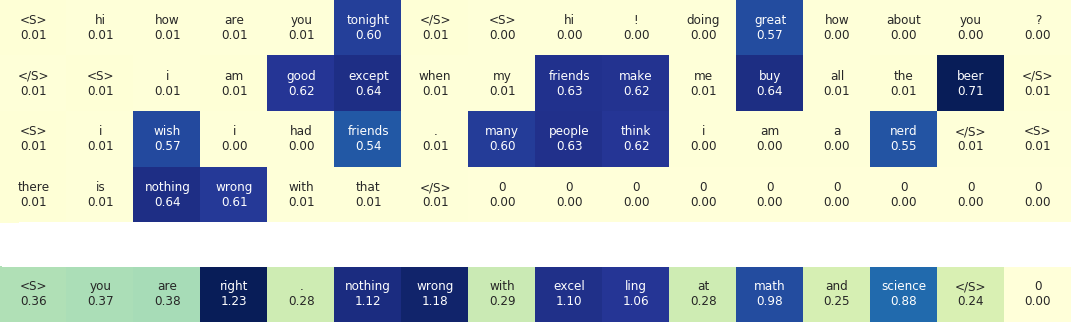}\label{fig:persona_vi2}}
\subfigure[Attentive dual encoder + WE + REG ]{
\includegraphics[width=.8\textwidth]{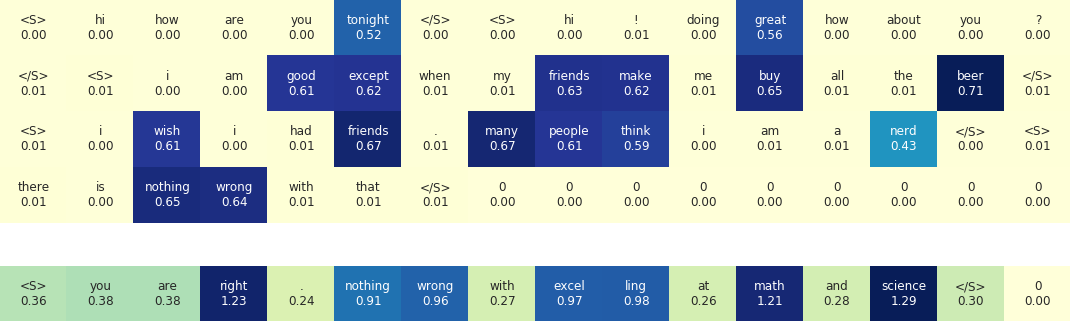}\label{fig:persona_vi3}}
\caption{Visualization on Persona dataset example with different competitors. (Top) Standard dual encoder model. (Middle) Standard dual encoder model with residual layer added for word embeddings. (Bottom) The attentive dual encoder model with all components included. \label{fig:persona_vi}}
\end{figure}

In addition to the quantitative results, model interpretability is also essential. We visualize the learned attention weights in Figure~\ref{fig:persona_vi}, where darker colors indicate higher weight values. Because of the entangled effect of word-level information from both the dialogue context and response encoders, the plain attentive dual encoder model cannot distinguish well the importance of different words. 
This can be observed by comparing Figure~\ref{fig:persona_vi1} and Figure~\ref{fig:persona_vi2}, where the former has similar weights for nearly all words. In contrast, \textbf{WE} connects the information between the raw word embedding and the deeply encoded features, providing more fine word-level interpretation for model prediction. 

The mutual information regularization term helps to alleviate the effect from uncorrelated words. As shown in Figure~\ref{fig:persona_vi3}, the emphasized predictive words are more reasonable and distinguished than the others. In the given example, one person is talking about being treated unfairly by his/her friends. From the human intuition, the attended words are expected to be 'buy beer', `nerd', `friends' in the dialogue context and `nothing wrong', `math', `science' in the response. For the method without mutual information regularization (Figure~\ref{fig:persona_vi2}), words like `excelling' are emphasized more than the others. This is caused by attentions on arbitrary words without any constraints. In summary, our two proposed components can help with improving interpretability for dialogue context-response prediction. A more rigorous study to quantify and evaluate the attention visualization effect will be done in the future.

\section{Conclusion}\label{sec:conclusion}
In this work, we presented a new interpretable model for dialogue response suggestion. The model is built upon the well-known dual encoder language model, where an attention mechanism is integrated to further improve the performance and show the word importance during the prediction. As a result, the proposed attentive dual encoder achieves better dialogue context-response prediction results on two datasets compared to existing methods. Additionally, we consider two problems to further improve the attention visualization quality. First, mutual information is used to constrain unimportant words in the dialogue context to have lower weights. Second, a residual layer is added between encoded sentence features and raw word embeddings, providing more fine-grained information on the word-level. 
With little effect on the prediction, the proposed methods further improve the word-level interpretability in the dialogue context-response prediction. 

\bibliographystyle{plainalt}
\bibliography{ref}

\begin{thebibliography}{10}

\bibitem{agrawal2018don}
A.~Agrawal, D.~Batra, D.~Parikh, and A.~Kembhavi.
\newblock Don't just assume; look and answer: Overcoming priors for visual
  question answering.
\newblock In {\em CVPR}, 2018.

\bibitem{anderson2018bottom}
P.~Anderson, X.~He, C.~Buehler, D.~Teney, M.~Johnson, S.~Gould, and L.~Zhang.
\newblock Bottom-up and top-down attention for image captioning and visual
  question answering.
\newblock In {\em CVPR}, 2018.

\bibitem{belghazi2018mine}
M.~I. Belghazi, A.~Baratin, S.~Rajeswar, S.~Ozair, Y.~Bengio, A.~Courville, and
  R.~D. Hjelm.
\newblock Mine: mutual information neural estimation.
\newblock {\em ICML}, 2018.

\bibitem{gino2019validity}
G.~Brunner, Y.~Liu, D.~Pascual, O.~Richter, and R.~Wattenhofer.
\newblock On the validity of self-attention as explanation in transformer
  models.
\newblock {\em arXiv preprint arXiv:1908.04211}, 2019.

\bibitem{donsker1975asymptotic}
M.~D. Donsker and S.~S. Varadhan.
\newblock Asymptotic evaluation of certain markov process expectations for
  large time, i.
\newblock {\em Communications on Pure and Applied Mathematics}, 1975.

\bibitem{faghri2017vse++}
F.~Faghri, D.~J. Fleet, J.~R. Kiros, and S.~Fidler.
\newblock Vse++: Improving visual-semantic embeddings with hard negatives.
\newblock {\em British Machine Vision Conference}, 2018.

\bibitem{gabrie2018entropy}
M.~Gabri{\'e}, A.~Manoel, C.~Luneau, N.~Macris, F.~Krzakala, L.~Zdeborov{\'a},
  et~al.
\newblock Entropy and mutual information in models of deep neural networks.
\newblock In {\em NeurIPS}, 2018.

\bibitem{ghazvininejad2018knowledge}
M.~Ghazvininejad, C.~Brockett, M.-W. Chang, B.~Dolan, J.~Gao, W.-t. Yih, and
  M.~Galley.
\newblock A knowledge-grounded neural conversation model.
\newblock In {\em AAAI}, 2018.

\bibitem{goodfellow2014generative}
I.~Goodfellow, J.~Pouget-Abadie, M.~Mirza, B.~Xu, D.~Warde-Farley, S.~Ozair,
  A.~Courville, and Y.~Bengio.
\newblock Generative adversarial nets.
\newblock In {\em NIPS}, 2014.

\bibitem{henderson2017efficient}
M.~Henderson, R.~Al-Rfou, B.~Strope, Y.-h. Sung, L.~Luk{\'a}cs, R.~Guo,
  S.~Kumar, B.~Miklos, and R.~Kurzweil.
\newblock Efficient natural language response suggestion for smart reply.
\newblock {\em arXiv preprint arXiv:1705.00652}, 2017.

\bibitem{hjelm2018learning}
R.~D. Hjelm, A.~Fedorov, S.~Lavoie-Marchildon, K.~Grewal, A.~Trischler, and
  Y.~Bengio.
\newblock Learning deep representations by mutual information estimation and
  maximization.
\newblock {\em ICLR}, 2019.

\bibitem{hsu2017unsupervised}
W.-N. Hsu, Y.~Zhang, and J.~Glass.
\newblock Unsupervised learning of disentangled and interpretable
  representations from sequential data.
\newblock In {\em NIPS}, 2017.

\bibitem{hu2017toward}
Z.~Hu, Z.~Yang, X.~Liang, R.~Salakhutdinov, and E.~P. Xing.
\newblock Toward controlled generation of text.
\newblock In {\em ICML}. JMLR. org, 2017.

\bibitem{jimenez2012soft}
S.~Jimenez, C.~Becerra, and A.~Gelbukh.
\newblock Soft cardinality: A parameterized similarity function for text
  comparison.
\newblock In {\em Proceedings of the First Joint Conference on Lexical and
  Computational Semantics}. Association for Computational Linguistics, 2012.

\bibitem{kingma2014adam}
D.~P. Kingma and J.~Ba.
\newblock Adam: A method for stochastic optimization.
\newblock {\em arXiv preprint arXiv:1412.6980}, 2014.

\bibitem{li2017targeting}
Y.~Li, K.~Dzirasa, L.~Carin, D.~E. Carlson, et~al.
\newblock Targeting eeg/lfp synchrony with neural nets.
\newblock In {\em NIPS}, 2017.

\bibitem{li2019way}
Y.~Li.
\newblock Which way are you going? imitative decision learning for path
  forecasting in dynamic scenes.
\newblock In {\em CVPR}, 2019.

\bibitem{liang2018focal}
J.~Liang, L.~Jiang, L.~Cao, L.-J. Li, and A.~G. Hauptmann.
\newblock Focal visual-text attention for visual question answering.
\newblock In {\em CVPR}, 2018.

\bibitem{lin2017structured}
Z.~Lin, M.~Feng, C.~N.~d. Santos, M.~Yu, B.~Xiang, B.~Zhou, and Y.~Bengio.
\newblock A structured self-attentive sentence embedding.
\newblock {\em arXiv preprint arXiv:1703.03130}, 2017.

\bibitem{lowe2015ubuntu}
R.~Lowe, N.~Pow, I.~Serban, and J.~Pineau.
\newblock The ubuntu dialogue corpus: A large dataset for research in
  unstructured multi-turn dialogue systems.
\newblock {\em SIGDIAL}, 2015.

\bibitem{mazare2018training}
P.-E. Mazar{\'e}, S.~Humeau, M.~Raison, and A.~Bordes.
\newblock Training millions of personalized dialogue agents.
\newblock {\em arXiv preprint arXiv:1809.01984}, 2018.

\bibitem{niu2019recursive}
Y.~Niu, H.~Zhang, M.~Zhang, J.~Zhang, Z.~Lu, and J.-R. Wen.
\newblock Recursive visual attention in visual dialog.
\newblock In {\em CVPR}, 2019.

\bibitem{palangi2018question}
H.~Palangi, P.~Smolensky, X.~He, and L.~Deng.
\newblock Question-answering with grammatically-interpretable representations.
\newblock In {\em Thirty-Second AAAI Conference on Artificial Intelligence},
  2018.

\bibitem{poole2019variational}
B.~Poole, S.~Ozair, A.~v.~d. Oord, A.~A. Alemi, and G.~Tucker.
\newblock On variational bounds of mutual information.
\newblock {\em ICML}, 2019.

\bibitem{qu2019weakly}
M.~Qu, J.~Tang, and Y.~Bengio.
\newblock Weakly-supervised knowledge graph alignment with adversarial
  learning.
\newblock {\em arXiv preprint arXiv:1907.03179}, 2019.

\bibitem{rocktaschel2015reasoning}
T.~Rockt{\"a}schel, E.~Grefenstette, K.~M. Hermann, T.~Ko{\v{c}}isk{\`y}, and
  P.~Blunsom.
\newblock Reasoning about entailment with neural attention.
\newblock {\em arXiv preprint arXiv:1509.06664}, 2015.

\bibitem{serrano2019attention_interpretable}
S.~Serrano and N.~A. Smith.
\newblock Is attention interpretable?
\newblock {\em arXiv preprint arXiv:1906.03731}, 2019.

\bibitem{shen2018disan}
T.~Shen, T.~Zhou, G.~Long, J.~Jiang, S.~Pan, and C.~Zhang.
\newblock Disan: Directional self-attention network for rnn/cnn-free language
  understanding.
\newblock In {\em Thirty-Second AAAI Conference on Artificial Intelligence},
  2018.

\bibitem{song2018towards}
Y.~Song, R.~Yan, Y.~Feng, Y.~Zhang, D.~Zhao, and M.~Zhang.
\newblock Towards a neural conversation model with diversity net using
  determinantal point processes.
\newblock In {\em AAAI}, 2018.

\bibitem{sordoni2015neural}
A.~Sordoni, M.~Galley, M.~Auli, C.~Brockett, Y.~Ji, M.~Mitchell, J.-Y. Nie,
  J.~Gao, and B.~Dolan.
\newblock A neural network approach to context-sensitive generation of
  conversational responses.
\newblock {\em arXiv preprint arXiv:1506.06714}, 2015.

\bibitem{sydorova2019interpretable}
A.~Sydorova, N.~Poerner, and B.~Roth.
\newblock Interpretable question answering on knowledge bases and text.
\newblock {\em arXiv preprint arXiv:1906.10924}, 2019.

\bibitem{trott2017interpretable}
A.~Trott, C.~Xiong, and R.~Socher.
\newblock Interpretable counting for visual question answering.
\newblock {\em ICLR}, 2018.

\bibitem{vaswani2017attention}
A.~Vaswani, N.~Shazeer, N.~Parmar, J.~Uszkoreit, L.~Jones, A.~N. Gomez,
  {\L}.~Kaiser, and I.~Polosukhin.
\newblock Attention is all you need.
\newblock In {\em NIPS}, 2017.

\bibitem{vedantam2019probabilistic}
R.~Vedantam, K.~Desai, S.~Lee, M.~Rohrbach, D.~Batra, and D.~Parikh.
\newblock Probabilistic neural-symbolic models for interpretable visual
  question answering.
\newblock {\em ICML}, 2019.

\bibitem{wallace2019universal}
E.~Wallace, S.~Feng, N.~Kandpal, M.~Gardner, and S.~Singh.
\newblock Universal adversarial triggers for attacking and analyzing nlp.
\newblock In {\em Proceedings of the 2019 Conference on Empirical Methods in
  Natural Language Processing and the 9th International Joint Conference on
  Natural Language Processing (EMNLP-IJCNLP)}, 2019.

\bibitem{wang2019multi}
X.~Wang, X.~Han, W.~Huang, D.~Dong, and M.~R. Scott.
\newblock Multi-similarity loss with general pair weighting for deep metric
  learning.
\newblock In {\em CVPR}, 2019.

\bibitem{wu2018starspace}
L.~Y. Wu, A.~Fisch, S.~Chopra, K.~Adams, A.~Bordes, and J.~Weston.
\newblock Starspace: Embed all the things!
\newblock In {\em AAAI}, 2018.

\bibitem{xu2017new}
A.~Xu, Z.~Liu, Y.~Guo, V.~Sinha, and R.~Akkiraju.
\newblock A new chatbot for customer service on social media.
\newblock In {\em Proceedings of the 2017 CHI Conference on Human Factors in
  Computing Systems}. ACM, 2017.

\bibitem{xu2019gated}
D.~Xu, J.~Ji, H.~Huang, H.~Deng, and W.-J. Li.
\newblock Gated group self-attention for answer selection.
\newblock {\em arXiv preprint arXiv:1905.10720}, 2019.

\bibitem{yan2018smarter}
R.~Yan and D.~Zhao.
\newblock Smarter response with proactive suggestion: A new generative neural
  conversation paradigm.
\newblock In {\em IJCAI}, 2018.

\bibitem{yang2018learning}
Y.~Yang, S.~Yuan, D.~Cer, S.-y. Kong, N.~Constant, P.~Pilar, H.~Ge, Y.-H. Sung,
  B.~Strope, and R.~Kurzweil.
\newblock Learning semantic textual similarity from conversations.
\newblock {\em arXiv preprint arXiv:1804.07754}, 2018.

\bibitem{young2013pomdp}
S.~Young, M.~Ga{\v{s}}i{\'c}, B.~Thomson, and J.~D. Williams.
\newblock Pomdp-based statistical spoken dialog systems: A review.
\newblock {\em Proceedings of the IEEE}, 2013.

\bibitem{yu2018qanet}
A.~W. Yu, D.~Dohan, M.-T. Luong, R.~Zhao, K.~Chen, M.~Norouzi, and Q.~V. Le.
\newblock Qanet: Combining local convolution with global self-attention for
  reading comprehension.
\newblock {\em ICLR}, 2018.

\bibitem{zhang2018addressee}
R.~Zhang, H.~Lee, L.~Polymenakos, and D.~Radev.
\newblock Addressee and response selection in multi-party conversations with
  speaker interaction rnns.
\newblock In {\em AAAI}, 2018.

\bibitem{zhang2018personalizing}
S.~Zhang, E.~Dinan, J.~Urbanek, A.~Szlam, D.~Kiela, and J.~Weston.
\newblock Personalizing dialogue agents: I have a dog, do you have pets too?
\newblock {\em ACL}, 2018.

\bibitem{zhao2016textual}
K.~Zhao, L.~Huang, and M.~Ma.
\newblock Textual entailment with structured attentions and composition.
\newblock In {\em Proceedings of COLING 2016, the 26th International Conference
  on Computational Linguistics}, 2016.

\bibitem{zhao2018unsupervised}
T.~Zhao, K.~Lee, and M.~Eskenazi.
\newblock Unsupervised discrete sentence representation learning for
  interpretable neural dialog generation.
\newblock {\em arXiv preprint arXiv:1804.08069}, 2018.

\bibitem{zhao2017learning}
T.~Zhao, R.~Zhao, and M.~Eskenazi.
\newblock Learning discourse-level diversity for neural dialog models using
  conditional variational autoencoders.
\newblock {\em ACL}, 2017.

\bibitem{zhou2018multi}
X.~Zhou, L.~Li, D.~Dong, Y.~Liu, Y.~Chen, W.~X. Zhao, D.~Yu, and H.~Wu.
\newblock Multi-turn response selection for chatbots with deep attention
  matching network.
\newblock In {\em ACL}, 2018.

\end{thebibliography}
\end{document}